%% file: main.tex
\documentclass[10pt,twocolumn,letterpaper]{article}

\usepackage[pagenumbers]{cvpr} 

\input{preamble}

\definecolor{cvprblue}{rgb}{0.21,0.49,0.74}
\usepackage[pagebackref,breaklinks,colorlinks,allcolors=cvprblue]{hyperref}

\def\paperID{25} 
\def\confName{CVPR}
\def\confYear{2026}

\title{PCM-NeRF: Probabilistic Camera Modeling for Neural Radiance Fields under Pose Uncertainty}

\author{Shravan Venkatraman*, Rakesh Raj Madavan*, Pavan Kumar Sathya Venkatesh* \\
    Mohamed bin Zayed University of AI, University of Amsterdam, University of Massachusetts Amherst \\
}

\begin{document}
\twocolumn[{%
    \renewcommand\twocolumn[1][]{#1}%
    \maketitle

    \begin{center}
\vspace{-25pt}
\small
\href{https://shravan-18.github.io/PCM-NeRF/}{%
    \faGlobe \hspace{4pt} \texttt{https://shravan-18.github.io/PCM-NeRF}
}
\vspace{6pt}
\end{center}
    \begin{center}
    \captionsetup{type=figure}
    \begin{minipage}{0.46\linewidth}
        \centering
        \includegraphics[width=\linewidth]{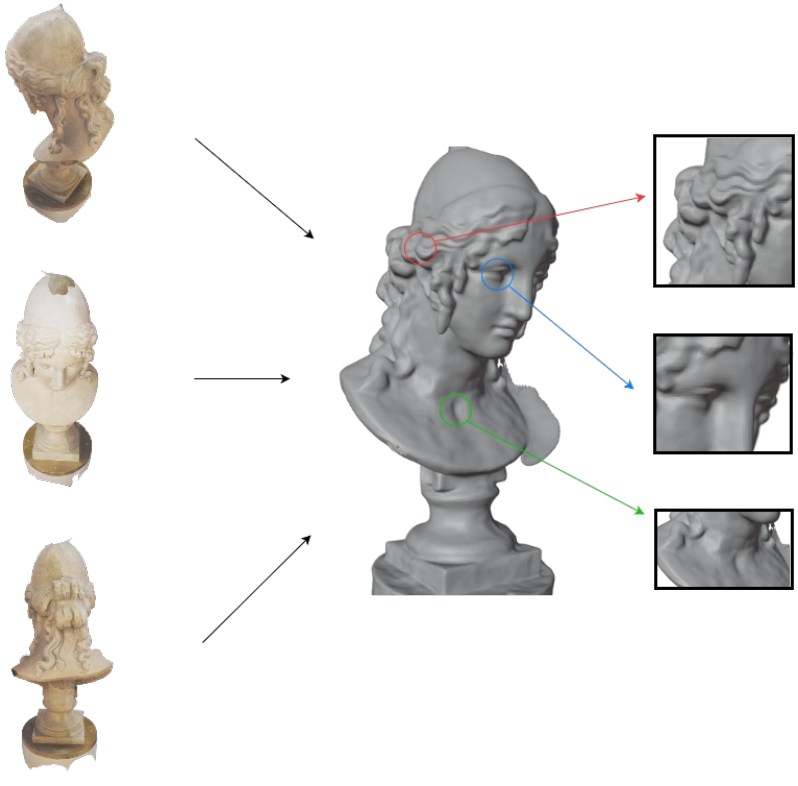}
    \end{minipage}%
    \hfill
    \begin{minipage}{0.53\linewidth}
        \centering
        \includegraphics[width=\linewidth]{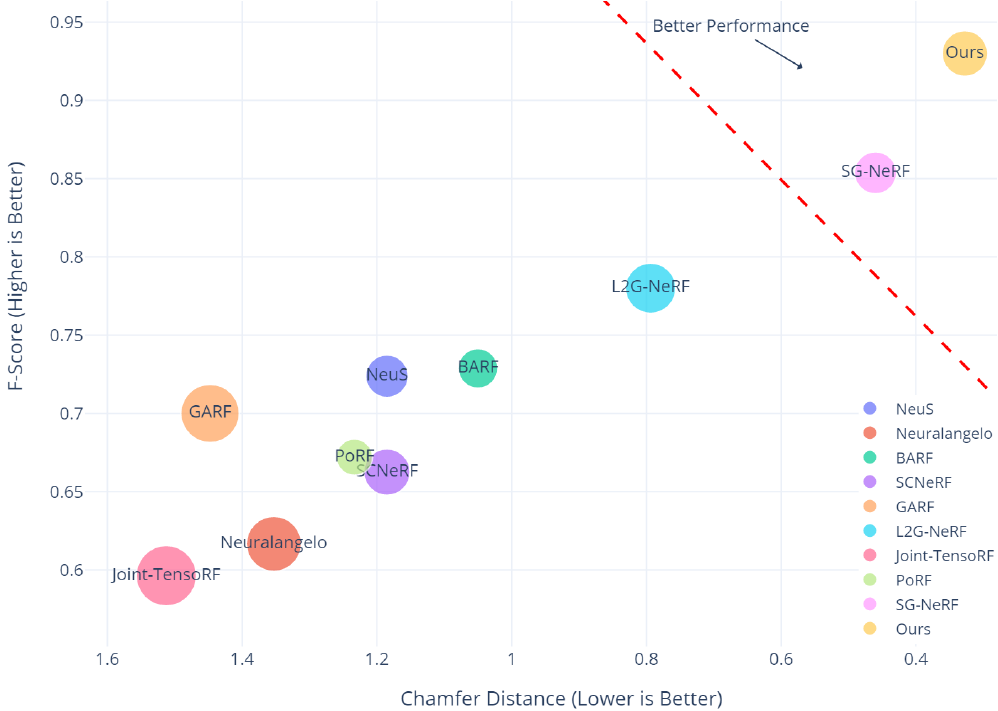}
    \end{minipage}
    \vspace{3pt}
    \caption{\textbf{Left:} Qualitative reconstruction showing PCM-NeRF's ability to capture fine surface detail. \textbf{Right:} PCM-NeRF outperforms all compared methods across both Chamfer Distance (lower is better) and F-Score (higher is better)}
    \label{fig:combined}
\end{center}
}]

{\renewcommand{\thefootnote}{*}\footnotetext{Equal contribution.}}

\input{sec/0_abstract}    
\input{sec/1_intro}
\input{sec/2_formatting}

\input{sec/3_finalcopy}

{
    \small
    \bibliographystyle{ieeenat_fullname}
    \bibliography{main}
}

\input{sec/X_suppl}

\end{document}

%% file: preamble.tex
\usepackage{graphicx}       
\usepackage{amsmath}        
\usepackage{amssymb}        
\usepackage{amsfonts}       
\usepackage{booktabs}       
\usepackage{multirow}       
\usepackage{booktabs}
\usepackage{subcaption}
\usepackage{float}
\usepackage{caption}
\usepackage{fontawesome5}
\usepackage{pifont}
\newcommand{\cmark}{\ding{51}}  
\newcommand{\xmark}{\ding{55}}  

\usepackage{xcolor}  
\usepackage[table]{xcolor}
\definecolor{mygray}{RGB}{220,220,220}  

%% file: sec/0_abstract.tex
\begin{abstract}
Neural surface reconstruction methods typically treat camera poses as fixed values, assuming perfect accuracy from Structure-from-Motion (SfM) systems. This assumption breaks down with imperfect pose estimates, leading to distorted or incomplete reconstructions. We present PCM-NeRF, a probabilistic framework that augments neural surface reconstruction with per-camera learnable uncertainty, built on top of SG-NeRF. Rather than treating all cameras equally throughout optimization, we represent each pose as a distribution with a learnable mean and variance, initialized from SfM correspondence quality. An uncertainty regularization loss couples the learned variance to view confidence, and the resulting uncertainty directly modulates the effective pose learning rate: uncertain cameras receive damped gradient updates, preventing poorly initialized views from corrupting the reconstruction. This lightweight mechanism requires no changes to the rendering pipeline and adds negligible overhead. Experiments on challenging scenes with severe pose outliers demonstrate that PCM-NeRF consistently outperforms state-of-the-art methods in both Chamfer Distance and F-Score, particularly for geometrically complex structures, without requiring foreground masks.
\end{abstract}

%% file: sec/1_intro.tex
\section{Introduction}
\label{sec:intro}

Neural surface reconstruction (NSR) has become a dominant approach for recovering detailed 3D geometry from 2D images, driven by advances in implicit representations and differentiable rendering~\cite{neusreference,dvr,nerf,idr,jignasu2024stitch,savani2024diffusing}. Methods such as IDR~\cite{idr} and NeuS~\cite{neusreference} leverage signed distance functions (SDFs) with volume rendering to reconstruct surfaces with high fidelity, with subsequent work extending this foundation through unified surface-volume rendering~\cite{unisurf} and multi-resolution hash grid encodings~\cite{neuralangelo}. Despite this progress, most approaches assume fixed and accurate camera poses, typically sourced from SfM pipelines~\cite{sfmrevisited,vggsfm,densesfm,light3rsfm}. In practice, however, SfM methods — even robust ones like COLMAP~\cite{sfmrevisited} — often fail or are not accurate under outlier captures, repetitive patterns, or occlusions, leading to inaccurate poses~\cite{Madhavan_2025_WACV}. Joint optimization strategies~\cite{scnerfreference,barfreference,bian2023nopenerf,chng2022gaussian,wang2021nerfminus,yan2023cfnerf} attempt to refine poses alongside scene geometry. Yet, these methods are generally limited to correcting small pose errors and fail when initialization contains large outliers~\cite{jointtensorf,truong2023sparf}. Severe pose inaccuracies often trap optimization in poor local minima, yielding distorted reconstructions, particularly in geometrically complex regions.

\begin{figure*}[t]
\centering
\includegraphics[width=\linewidth]{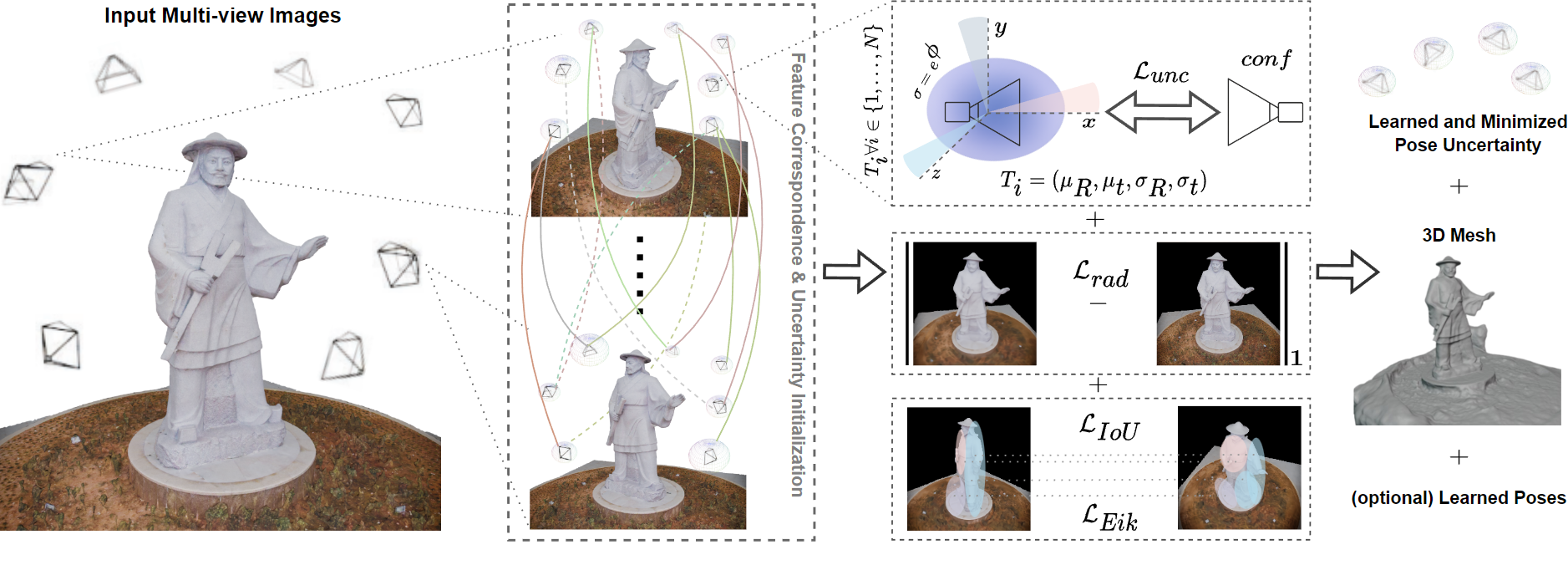}
\caption{Overview of our PCM-NeRF pipeline. From left to right: (1) Multi-view images are processed to establish feature correspondences and initialize pose uncertainty. (2) Our probabilistic camera model represents each pose as a distribution with learnable parameters. (3) Joint optimization with uncertainty-guided regularization, photometric rendering losses, and volumetric distribution alignment produces high-quality surface reconstruction while learning pose uncertainty. The framework can also optionally optimize camera poses when needed.}
\label{fig:pipeline}
\end{figure*}

Recent methods have begun addressing neural reconstruction under unreliable camera poses. SG-NeRF~\cite{sgnerf} uses scene graph optimization with adaptive confidence estimation to reduce the impact of pose outliers, while L2G-NeRF~\cite{chen2023localglobal} adopts a local-to-global registration strategy to correct poor initializations, and Joint-TensoRF~\cite{jointtensorf} integrates pose refinement into a factorized tensor representation for efficient optimization. However, these approaches treat camera poses as deterministic, limiting their ability to model uncertainty under large pose deviations where optimization often fails to escape poor local minima — a gap that uncertainty modeling in implicit neural rendering~\cite{nerfotg,ucnerf,bayesrays,bayesiannerf,stochasticnerf} has yet to bridge for NSR. To this end, we propose PCM-NeRF, a probabilistic framework built directly on SG-NeRF~\cite{sgnerf} that represents each pose as a multivariate normal distribution with trainable mean and covariance, capturing varying confidence in both translation and rotation. The learned uncertainty directly modulates the effective pose learning rate, damping updates for poorly constrained cameras and preventing outlier views from destabilizing reconstruction — while inheriting the scene graph structure, feature correspondence pipeline, and volumetric consistency loss from SG-NeRF~\cite{sgnerf} unchanged.

Figure~\ref{fig:pipeline} presents an overview of our PCM-NeRF framework. Starting from multi-view images, the method initializes a feature correspondence structure and estimates pose uncertainty by evaluating cross-view matching quality. Each camera pose $T_i$ is modeled as a distribution with learnable means $(\mu_R, \mu_t)$ and associated uncertainties $(\sigma_R, \sigma_t)$ for rotation and translation, coupled with view confidence scores computed from correspondence density and rendering quality and optimized via $\mathcal{L}_{\text{unc}}$. The training objective combines $\mathcal{L}_{\text{rad}}$, $\mathcal{L}_{\text{Eik}}$, and $\mathcal{L}_{\text{IoU}}$ inherited from SG-NeRF~\cite{sgnerf} to promote geometric consistency across views. Rather than explicitly re-weighting views during rendering, uncertainty parameters act as a regularization prior, guiding optimization toward reliable pose estimates and improving surface reconstruction regardless of whether poses are fixed or jointly optimized.

Our contributions can be summarized as follows:
\begin{itemize}
    \item We introduce a probabilistic per-camera pose representation for NSR in which each pose is modeled as a distribution in SE(3) with a learnable mean and per-axis variance, initialized from SfM correspondence quality and evolved jointly with the scene.
    \item We derive an uncertainty-modulated pose learning rate  that inversely scales each camera's effective gradient step by its learned uncertainty, automatically preventing outlier views from destabilizing reconstruction without manual filtering, threshold selection, or any changes to the rendering pipeline.
    \item We propose a cross-modal consistency loss that aligns learned variance parameters with view confidence derived from both geometric correspondences and rendering quality, anchoring uncertainty to observable evidence rather than rendering gradients alone.
    \item We demonstrate state-of-the-art surface reconstruction on challenging outlier-pose scenes, achieving the lowest Chamfer Distance on 6 of 8 scenes and the highest F-Score on 6 of 8 scenes, outperforming all compared baselines.
\end{itemize}

%% file: sec/2_formatting.tex
\section{Related Work}
\label{sec:relatedwork}

\noindent\textbf{Neural Surface Reconstruction.} Neural surface reconstruction (NSR) methods recover 3D geometry from multi-view images using implicit representations. Foundational methods such as NeuS~\cite{neusreference}, VolSDF~\cite{volsdf}, and IDR~\cite{idr} established SDF-based volume rendering as the dominant paradigm, with UNISURF~\cite{unisurf} unifying radiance fields and implicit surfaces for joint appearance-geometry optimization. Subsequent work has pushed this foundation in several directions: high-frequency detail recovery~\cite{hfneus,dneus}, high-fidelity reconstruction via multi-resolution hash grids~\cite{neuralangelo}, arbitrary-topology surfaces using unsigned distance fields~\cite{neat,neuraludf}, and efficiency through permutohedral lattices or voxel grids~\cite{permutosdf,voxurf}. Further extensions have targeted tri-plane feature encodings~\cite{petneus}, shadow ray supervision for lighting robustness~\cite{shadowneus}, and kernel-free point cloud serialization~\cite{noksr}. Despite this breadth, all of the above assume accurate input poses, leaving reconstruction quality vulnerable to SfM failures and outlier cameras.

\noindent\textbf{Camera Pose Estimation.} Joint optimization of camera poses and scene representation has been a focal point in neural rendering research. BARF~\cite{barfreference} introduced frequency-filtered bundle adjustment to stabilize training under inaccurate poses. Subsequent methods approach this from varied angles: synchronizing local unposed NeRFs~\cite{lunerf}, combining keypoint-based optimization with renderable neural codes~\cite{nerfels}, Monte Carlo sampling with robust losses for 6-DoF estimation~\cite{parallelinversion}, feature-integrated reprojection error formulations~\cite{neuralreprojection}, and joint optimization within multi-resolution hash encoding frameworks~\cite{heo2023robustcameraposerefinement}. However, all treat camera poses as deterministic point estimates, providing no mechanism to quantify or leverage per-camera reliability during training.

\noindent\textbf{Uncertainty in Neural Rendering.} For uncertainty modeling, NVINS~\cite{nvins} coupled NeRF-based localization with visual-inertial odometry and uncertainty estimation; VERF~\cite{verf} added runtime pose assurance through confidence-aware NeRF rendering. Uncertainty sources \cite{uncertaintysources} analyzed and estimated various uncertainty types in NeRF and Gaussian Splatting. Bayesian Relocalization~\cite{bayesianrelocalization} used Bayesian CNNs for real-time monocular pose estimation with uncertainty quantification. Uncertainty-aware PnP~\cite{uncertaintypnp} improved pose estimation by incorporating uncertainty in both 2D and 3D features. While these works demonstrate the value of uncertainty in neural rendering pipelines, they focus on appearance uncertainty or post-hoc pose assurance rather than integrating per-camera pose uncertainty directly into NSR optimization — the gap that PCM-NeRF addresses.

\section{Method}
\label{sec:method}

\subsection{Background}

\noindent\textbf{Neural Implicit Surface Representation.} We represent a surface $\mathcal{S}$ as the zero-level set of a SDF $f: \mathbb{R}^3 \rightarrow \mathbb{R}$ parameterized by a neural network: $\mathcal{S} = \{x \in \mathbb{R}^3 | f(x) = 0\}$. The SDF $f(x)$ outputs the signed distance from point $x$ to the nearest surface point, with negative values inside the surface and positive values outside. Additionally, a color field $c: \mathbb{R}^3 \times \mathbb{S}^2 \rightarrow \mathbb{R}^3$ maps a spatial position $x$ and viewing direction $v$ to an RGB color.

\noindent\textbf{Volume Rendering and Neural Surface Reconstruction.} To render a ray $r(t) = o + tv$ originating from camera center $o$ with direction $v$, NeuS \cite{neusreference} introduces a density function derived from the SDF that enables volume rendering while ensuring accurate surface reconstruction. The key insight is to define an S-density $\phi_s(f(x))$ from the SDF, where $\phi_s$ is the derivative of the sigmoid function $\Phi_s$: $\phi_s(x) = \frac{se^{-sx}}{(1+e^{-sx})^2}$.

The rendering integral for a pixel's color $C(r)$ is:
\(
C(r) = \int_{0}^{\infty} w(t)c(r(t),v)dt
\)
where $w(t) = T(t)\rho(t)$ is a weight function, $T(t) = \exp(-\int_{0}^{t}\rho(u)du)$ is the accumulated transmittance, and $\rho(t)$ is an opaque density function derived from the S-density. In practice, this integral is approximated by sampling points along the ray and using a discrete formulation: $\hat{C}(r) = \sum_{i=1}^{n} T_i \alpha_i c_i$, where $T_i = \prod_{j=1}^{i-1}(1-\alpha_j)$ and $\alpha_i$ is the discrete opacity. This enables differentiable rendering while ensuring precise surface reconstruction by linking density estimation to the SDF, effectively capturing surface boundaries during optimization.

\noindent\textbf{Camera Model in Neural Rendering.} Conventional neural rendering approaches assume fixed camera poses $P_i = (R_i, t_i)$, where $R_i \in SO(3)$ is a rotation matrix and $t_i \in \mathbb{R}^3$ is a translation vector. These poses are typically obtained from SfM systems and treated as ground truth without accounting for uncertainty. This treatment can lead to suboptimal reconstruction results, especially with erroneous pose estimates or scenes with sudden depth changes.

\subsection{Probabilistic Camera Pose Representation}

We develop a principled approach that explicitly models camera poses as probability distributions with learnable parameters, enhancing how neural rendering systems utilize pose information.

\begin{figure}[t]
  \centering
  \includegraphics[width=\linewidth]{assets/PCM-NeRF_1.drawio.pdf}
  \caption{\textbf{Volumetric Distribution Alignment (VDA) for geometric consistency.} 
  Before alignment (left), initial camera pose uncertainty ($\sigma_{R}$, $\sigma_{t}$) causes intersecting rays from views $C_{i}$ and $C_{j}$ to produce spatially separated density blobs along the ray distance $t$. 
  Our VDA optimization (right) utilizes the volumetric consistency loss to align top-$k$ density peaks. 
  By representing density as a Mixture of Gaussians (MoG), the method maximizes volumetric overlap on the shared surface while simultaneously minimizing learnable pose uncertainty parameters.}
  \label{fig:vda_alignment}
\end{figure}

\begin{figure*}[t]
  \centering
  \includegraphics[width=0.9\linewidth,height=0.35\textheight]{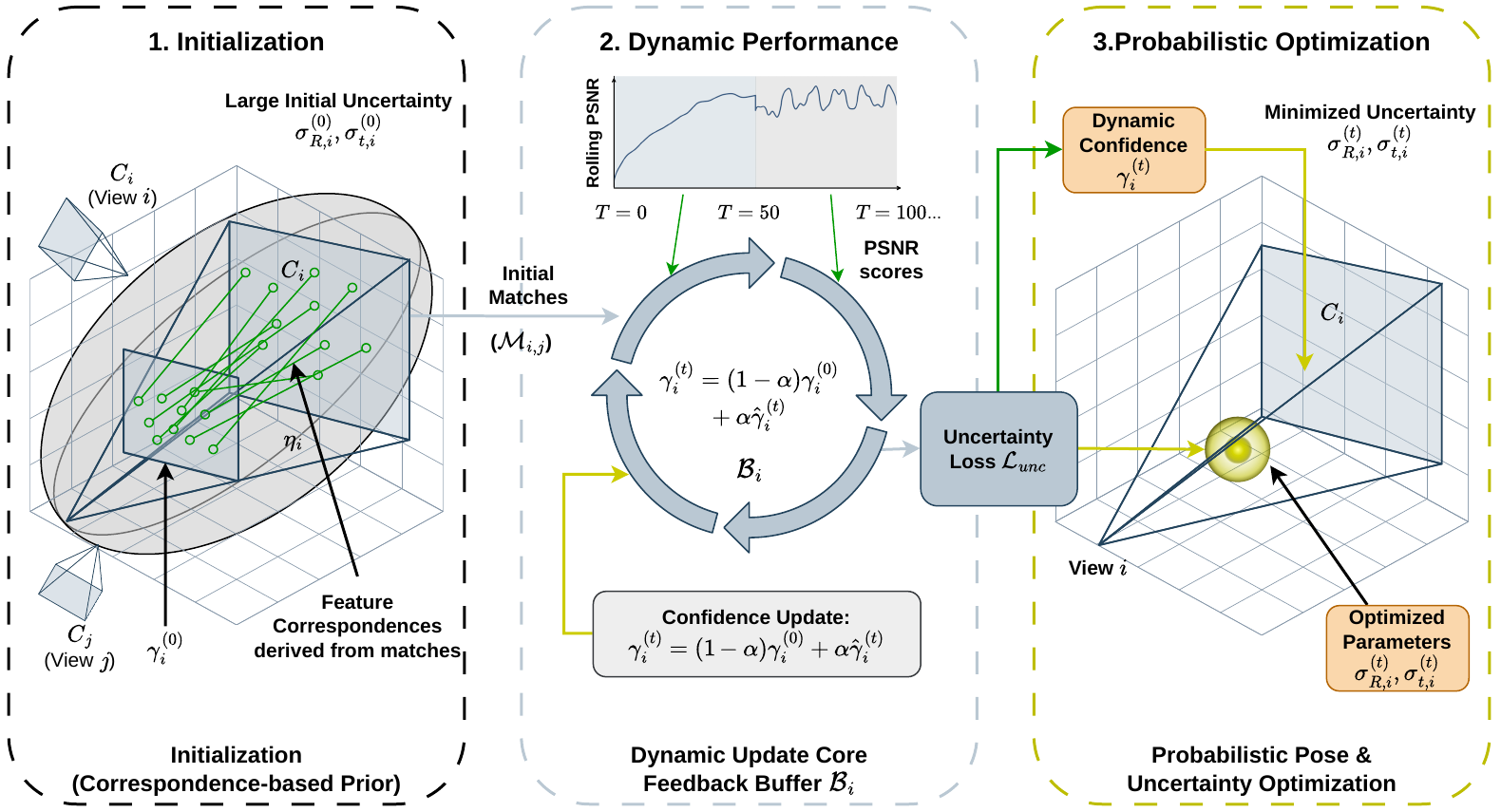}
  \caption{\textbf{Overview of the Uncertainty Feedback Loop.} 
  The framework establishes a dynamic link between view reliability and pose precision. 
  (1) \textbf{Initialization}: Initial reliability is computed from feature correspondences $\mathcal{M}_{i,j}$ to provide a starting prior for camera uncertainty. 
  (2) \textbf{Dynamic Performance}: A feedback buffer $\mathcal{B}_{i}$ tracks rolling PSNR scores to compute the dynamic confidence $\gamma_{i}^{(t)} = (1-\alpha)\gamma_{i}^{(0)} + \alpha\hat{\gamma}_{i}^{(t)}$. 
  (3) \textbf{Probabilistic Optimization}: The uncertainty loss $\mathcal{L}_{unc}$ regularizes the learnable parameters $\sigma_{R,i}, \sigma_{t,i}$, which in turn modulates the effective pose learning rate, resulting in minimized uncertainty and robust surface reconstruction even in the presence of initial pose outliers.}
  \label{fig:uncertainty_loop}
\end{figure*}

Our approach is based on the observation that camera pose uncertainty varies across different views within a scene. Views with repetitive patterns, textureless regions, or unusual viewpoints typically have higher pose uncertainty $\sigma^2_{\text{high}}$, while views with distinctive features and good correspondence have lower uncertainty $\sigma^2_{\text{low}} \ll \sigma^2_{\text{high}}$. By modeling this uncertainty $\sigma^2_i$ explicitly for each view $i$, our framework can adaptively weight the contribution of different views in the optimization process.

Each camera pose $P_i$ is modeled as a distribution in $SE(3)$, with mean $\mu_i = (r_i, t_i)$ and diagonal covariance $\Sigma_i = \text{diag}(\sigma_{r,i}^2, \sigma_{t,i}^2)$:
$$
P_i \sim \mathcal{N}(\mu_i, \Sigma_i)
$$

Here, $r_i \in \mathbb{R}^3$ is the rotation in axis-angle form, and $t_i \in \mathbb{R}^3$ is the translation. Both $(r_i, t_i)$ and the uncertainty parameters $(\sigma_{r,i}^2, \sigma_{t,i}^2)$ are learned. Initial mean poses come from SfM; uncertainties are initialized based on view reliability (see Section~3.3, Further details are presented in the supplementary). To ensure positive variances and stability, we use log-space parameterization:
$$
\sigma_{r,i}^2 = \exp(\lambda_{r,i}) 
$$$$
\sigma_{t,i}^2 = \exp(\lambda_{t,i})
$$

The variance parameters $\lambda_{r,i}$ and $\lambda_{t,i}$ serve two 
coupled purposes despite not entering the rendering forward pass directly. 
First, they act as a regularization prior anchored to observable evidence 
--- the correspondence quality and rendering fidelity of each view --- 
ensuring the learned uncertainty reflects genuine pose reliability. Second, 
and more operationally, they directly control the effective pose learning 
rate: a camera whose learned uncertainty is high has its pose gradient 
damped, preventing it from corrupting the reconstruction even if its SfM 
initialization is poor. While a full probabilistic rendering integral over the pose distribution is theoretically desirable:
$
\mathbb{E}[C(r)] = \int_{P_i} C(r|P_i) \cdot p(P_i|\mu_i, \Sigma_i) \, dP_i
$
evaluating this exactly is computationally prohibitive in the inner training loop. Instead, we use the mean pose $\mu_i$ for all rendering operations — the variance parameters $\sigma_{r,i}^2$ and $\sigma_{t,i}^2$ are not used in the forward rendering pass (additional details are presented in the supplementary). Their role is as a learned regularization prior: shaped by the uncertainty loss (Section~3.3.2), they in turn modulate the effective pose learning rate during optimization. Specifically, the learning rate for camera $i$ is scaled as:
$$
\eta_i^{\text{pose}} = \frac{\eta_0}{1 + \bar{\sigma}_i \cdot \kappa}
$$
where $\bar{\sigma}_i = (\sigma_{r,i}^2 + \sigma_{t,i}^2)/6$ is the mean scalar uncertainty magnitude and $\kappa$ controls sensitivity. Cameras with high uncertainty therefore take smaller pose gradient steps, preventing poorly initialized views from destabilizing the reconstruction (implementation details regarding per-camera gradient damping mechanism are presented in the supplementary). This is the primary mechanism through which uncertainty improves optimization, and it keeps inference-time cost identical to standard deterministic pose refinement. The full uncertainty feedback loop is illustrated in Figure~\ref{fig:uncertainty_loop}.

\subsection{Uncertainty-Guided Optimization}

Building on our probabilistic camera pose representation, we develop an uncertainty-guided optimization framework that dynamically assesses view reliability and couples it with uncertainty parameters during training. This approach builds directly on scene graph-based optimization~\cite{sgnerf}: we inherit the feature correspondence pipeline, dynamic confidence update, and adaptive ray sampling from SG-NeRF~\cite{sgnerf} unchanged, and extend them with the probabilistic uncertainty coupling described below.

\subsubsection{View Reliability Assessment}
Our framework begins by estimating the reliability of each camera view 
based on multi-view correspondence structure. For a set of images 
$\mathcal{I} = \{I_i\}_{i=1}^N$, we compute an initial reliability 
score $\eta_i = \frac{\sum_{j \in \mathcal{N}_i} |\mathcal{M}_{i,j}|}
{|\mathcal{N}_i|}$ for each view $i$, where $\mathcal{N}_i$ is the set 
of neighboring views, $\mathcal{M}_{i,j}$ is the set of matched 
keypoints between views $i$ and $j$, and $|\cdot|$ denotes set 
cardinality. These scores are normalized to obtain initial confidence 
values $\gamma_i^{(0)} = (\eta_i - \min_j \eta_j)/(\max_j \eta_j - 
\min_j \eta_j + \epsilon)$, where $\epsilon = 10^{-5}$ prevents 
division by zero.

During training, we maintain a buffer $\mathcal{B}_i$ of recent PSNR 
scores for each view and update confidence as $\gamma_i^{(t)} = 
(1-\alpha) \cdot \gamma_i^{(0)} + \alpha \cdot \hat{\gamma}_i^{(t)}$, 
where $\hat{\gamma}_i^{(t)}$ is the normalized PSNR history and 
$\alpha$ controls the balance between correspondence-based and 
rendering-based estimates (Full details are given in the supplementary). This creates a feedback mechanism where 
views that render well gain higher confidence while poorly rendering 
views are downweighted. Note that the correspondence-based 
initialization $\eta_i$, the dynamic PSNR confidence update, and the 
adaptive ray sampling that prioritizes high-confidence views are all 
adopted from SG-NeRF~\cite{sgnerf}; our contribution is the coupling 
of these confidence values to per-camera uncertainty parameters, 
described next.

\subsubsection{Uncertainty Regularization Loss}
We propose an uncertainty regularization loss that couples the view 
confidence values $\gamma_i$ with the uncertainty parameters 
$\boldsymbol{\sigma}_{r,i}^2 \in \mathbb{R}^3$ and 
$\boldsymbol{\sigma}_{t,i}^2 \in \mathbb{R}^3$, effectively encouraging 
consistency between the predicted confidences and the associated 
uncertainties:
$$
\mathcal{L}_{\text{uncertainty}} = \frac{1}{N} \sum_{i=1}^N \left| 
\left( \frac{\|\boldsymbol{\sigma}_{r,i}^2\|_1 + 
\|\boldsymbol{\sigma}_{t,i}^2\|_1}{6} \right) - (1 - \gamma_i) \right|
$$
This loss acts as a cross-modal consistency constraint between two 
partially independent signals: the confidence $\gamma_i$, which is 
grounded in SfM correspondence quality at initialization and updated 
via rendering feedback, and the learned variance parameters 
$\boldsymbol{\sigma}_{r,i}^2$, $\boldsymbol{\sigma}_{t,i}^2$, which 
are shaped by gradient flow from the reconstruction. Importantly, 
while these two signals are driven into alignment by the loss, they 
remain distinct: $\gamma_i$ is derived from external geometric evidence 
--- SfM match density and rendering PSNR --- and is not back-propagated 
through the neural rendering, whereas $\boldsymbol{\sigma}_i^2$ is a 
free parameter updated via gradient descent, allowing it to settle at 
a value consistent with both the geometric prior and the evolving 
reconstruction quality rather than being fixed at initialization. The 
loss bridges these two signals without collapsing them. By driving them 
into alignment, it ensures that the learned uncertainty is anchored to 
geometric evidence rather than determined by rendering gradients alone. 
The consequence is direct and practical: through the learning rate 
modulation described in Section~3.2, high-confidence views maintain 
aggressive pose updates while uncertain views are automatically damped, 
as illustrated in Figure~\ref{fig:uncertainty_loop}.

\subsection{Geometric Consistency through Distribution Alignment}
We adopt the volumetric distribution alignment approach of 
SG-NeRF~\cite{sgnerf} for geometric consistency. For views $i$ and 
$j$ with feature correspondences $\mathcal{M}_{i,j}$, we cast rays 
$r_i(t) = o_i + td_i$ and $r_j(t) = o_j + td_j$ through corresponding 
keypoints. Each ray's volume rendering produces a density distribution 
represented as a weighted point cloud $\{(p_k, w_k)\}$, where $p_k$ 
are 3D points and $w_k$ are their weights. We quantify alignment using 
an IoU loss $\mathcal{L}_{\text{IoU}} = 1 - \frac{\sum_v \min(D_i(v), 
D_j(v))}{\sum_v \max(D_i(v), D_j(v))}$, where $D_i(v)$ and $D_j(v)$ 
are voxelized density distributions projected onto a discretized volume 
grid with resolution $R^3$ ($R=64$), using only the top-$k$ 
highest-weight points ($k=8$) for efficiency. Figure~\ref{fig:vda_alignment} 
illustrates how this loss aligns density distributions that are initially 
separated due to pose uncertainty, driving them toward a common surface 
as optimization proceeds. Full implementation details of the Mixture-of-Gaussians construction 
and voxel grid setup are provided in the supplementary.

Our complete training objective is $\mathcal{L} = \mathcal{L}_{\text{color}} 
+ \lambda_{\text{eik}}\mathcal{L}_{\text{eik}} + \lambda_{\text{IoU}}
\mathcal{L}_{\text{IoU}} + \lambda_{\text{unc}}\mathcal{L}_{\text{uncertainty}}$, 
where $\mathcal{L}_{\text{color}}$ is the L1 color reconstruction loss 
between rendered and ground truth pixels, $\mathcal{L}_{\text{eik}}$ is 
the Eikonal regularization enforcing $\|\nabla f(x)\|=1$ for SDF 
properties, and $\lambda$ terms are weighting coefficients 
($\lambda_{\text{eik}}=0.1$, $\lambda_{\text{IoU}}=0.2$, 
$\lambda_{\text{unc}}=0.05$). The uncertainty term 
$\mathcal{L}_{\text{uncertainty}}$ is introduced gradually using a 
warm-up scheme after $T_{\text{warmup}}$ iterations, ramping up over 
$T_{\text{rampup}}$ iterations. The adaptive ray sampling and 
coarse-to-fine strategy with Gaussian blurring are both inherited from 
SG-NeRF~\cite{sgnerf} and used unchanged; together with our uncertainty 
module, they help avoid local minima when poses are unreliable.

%% file: sec/3_finalcopy.tex
\section{Experiments and Results}
Extensive experiments were conducted to evaluate the robustness of PCM-NeRF. We utilize the challenging inward-facing (developed from BlendedMVS\cite{yao2020blendedmvslargescaledatasetgeneralized}) dataset with camera outliers proposed by \cite{sgnerf}, which features varying degrees of pose uncertainty and outliers. We detail our experimental setup and present comprehensive comparisons against state-of-the-art neural surface reconstruction methods.

\begin{figure}[t]
  \centering
  \includegraphics[width=0.8\linewidth]{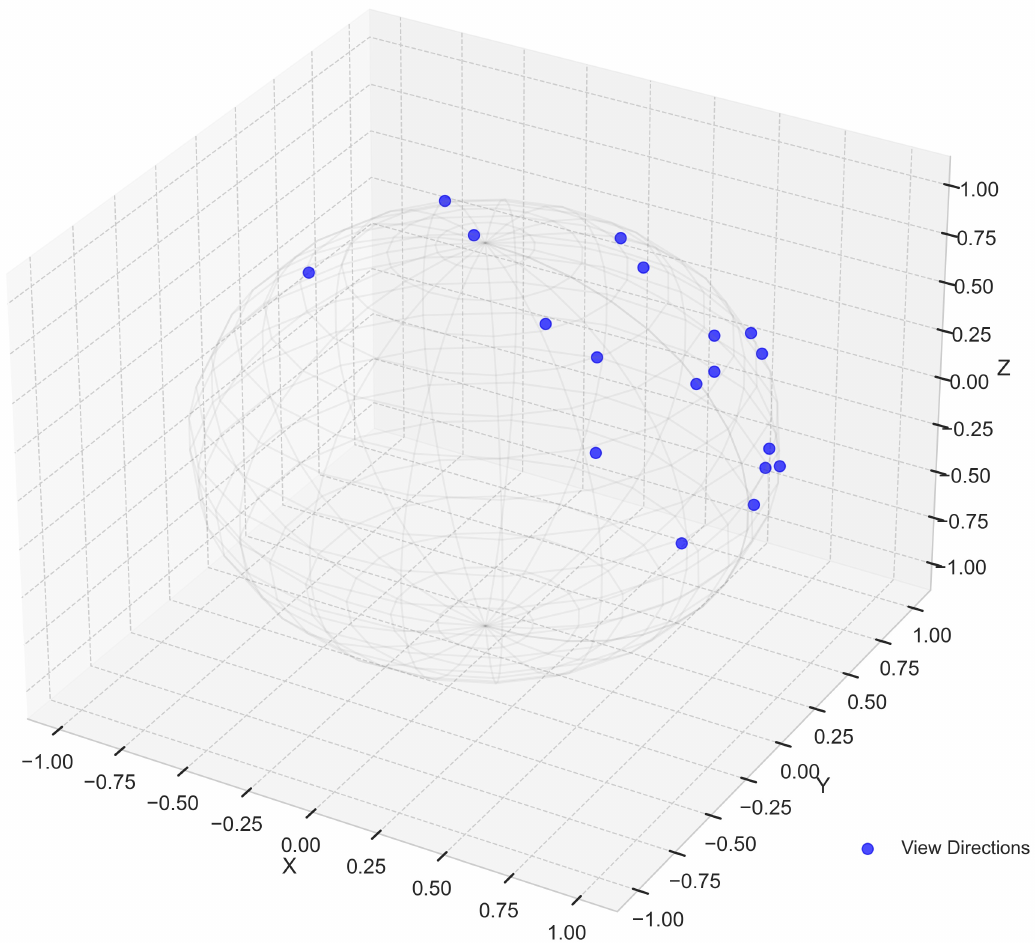}
  \caption{\textbf{Camera Viewpoint Distribution.} 
  A 3D visualization of the camera positions for a representative scene (e.g., 'Farmer') after outlier pose removal. Each blue point represents a camera origin $O_i$ in $SE(3)$, illustrating the dense hemispherical coverage required for capturing complex surface geometry in our inward-facing dataset.}
  \label{fig:viewpoints}
\end{figure}

\begin{figure*}[t]
\centering
\begin{minipage}[t]{0.62\linewidth}
\centering
\captionof{table}{\small Comparison of Different Methods Across Scenes on Chamfer distance. Lower values indicate better geometric accuracy. Our method achieves the lowest Chamfer distance on six out of eight scenes, outperforming both implicit and explicit neural reconstruction baselines across diverse object categories.}
\label{tab:comparison}
\footnotesize
\setlength{\tabcolsep}{3pt}
\begin{tabular}{@{}l@{\hspace{3pt}}|@{\hspace{3pt}}cccccccc@{}}
\toprule
\textbf{Method} & \textbf{Baby} & \textbf{Bear} & \textbf{Bell} & \textbf{Clock} & \textbf{Deaf} & \textbf{Farmer} & \textbf{Pavilion} & \textbf{Sculpture} \\
\midrule
NeuS~\cite{neusreference} & 0.69 & 0.31 & 3.33 & 1.16 & 0.55 & 2.49 & 0.29 & 0.66 \\
Neuralangelo~\cite{neuralangelo} & 0.70 & 0.65 & — & 0.38 & 0.59 & 4.89 & 1.95 & 0.31 \\
BARF~\cite{barfreference} & 1.08 & 0.28 & 3.31 & 0.19 & 0.46 & 2.13 & 0.38 & 0.57 \\
SCNeRF~\cite{scnerfreference} & 1.19 & 0.27 & 3.74 & 1.33 & 0.46 & 1.45 & 0.23 & 0.81 \\
GARF~\cite{chng2022gaussian} & 2.04 & 2.25 & 3.09 & 0.50 & 0.59 & 1.58 & 0.96 & 0.57 \\
L2G-NeRF~\cite{chen2023localglobal} & 1.15 & 0.29 & 1.26 & 0.24 & 0.40 & 2.18 & 0.46 & 0.37 \\
Joint-TensoRF~\cite{jointtensorf} & 3.11 & 1.22 & 2.49 & 0.36 & 0.36 & 2.51 & 1.35 & 0.70 \\
PoRF~\cite{bian2024porfposeresidualfield} & 0.31 & 0.49 & — & 0.30 & 0.30 & 3.80 & 2.20 & — \\
SG-NeRF~\cite{sgnerf} & 0.56 & 0.25 & 0.98 & \textbf{0.15} & 0.45 & 0.87 & \textbf{0.20} & 0.22 \\
\midrule
\textbf{Ours} & \textbf{0.25} & \textbf{0.24} & \textbf{0.51} & 0.16 & \textbf{0.23} & \textbf{0.79} & 0.24 & \textbf{0.20} \\
\bottomrule
\end{tabular}
\end{minipage}
\hfill
\begin{minipage}[t]{0.35\linewidth}
\vspace*{-0.5\baselineskip}
\centering
\includegraphics[width=\linewidth]{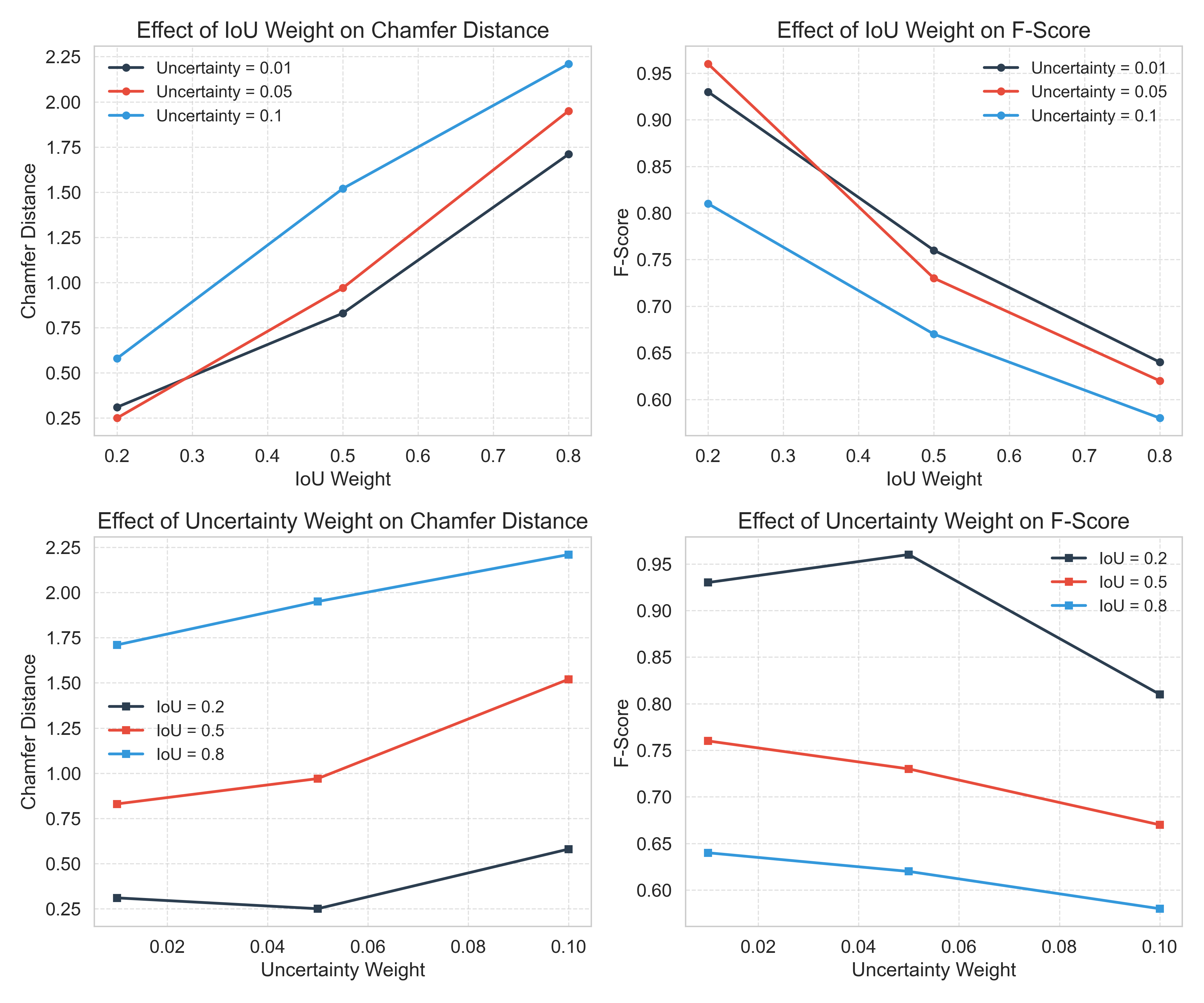}
\vspace{-23pt}
\captionof{figure}{\small Effect of weights on Chamfer Distance and F-Score. Lower IoU weights combined with moderate uncertainty weights consistently yield better reconstruction performance.}
\label{fig:weight_effects}
\end{minipage}
\end{figure*}

\begin{figure*}[t]
    \centering
    
    \includegraphics[width=1\textwidth, height=0.35\textheight]{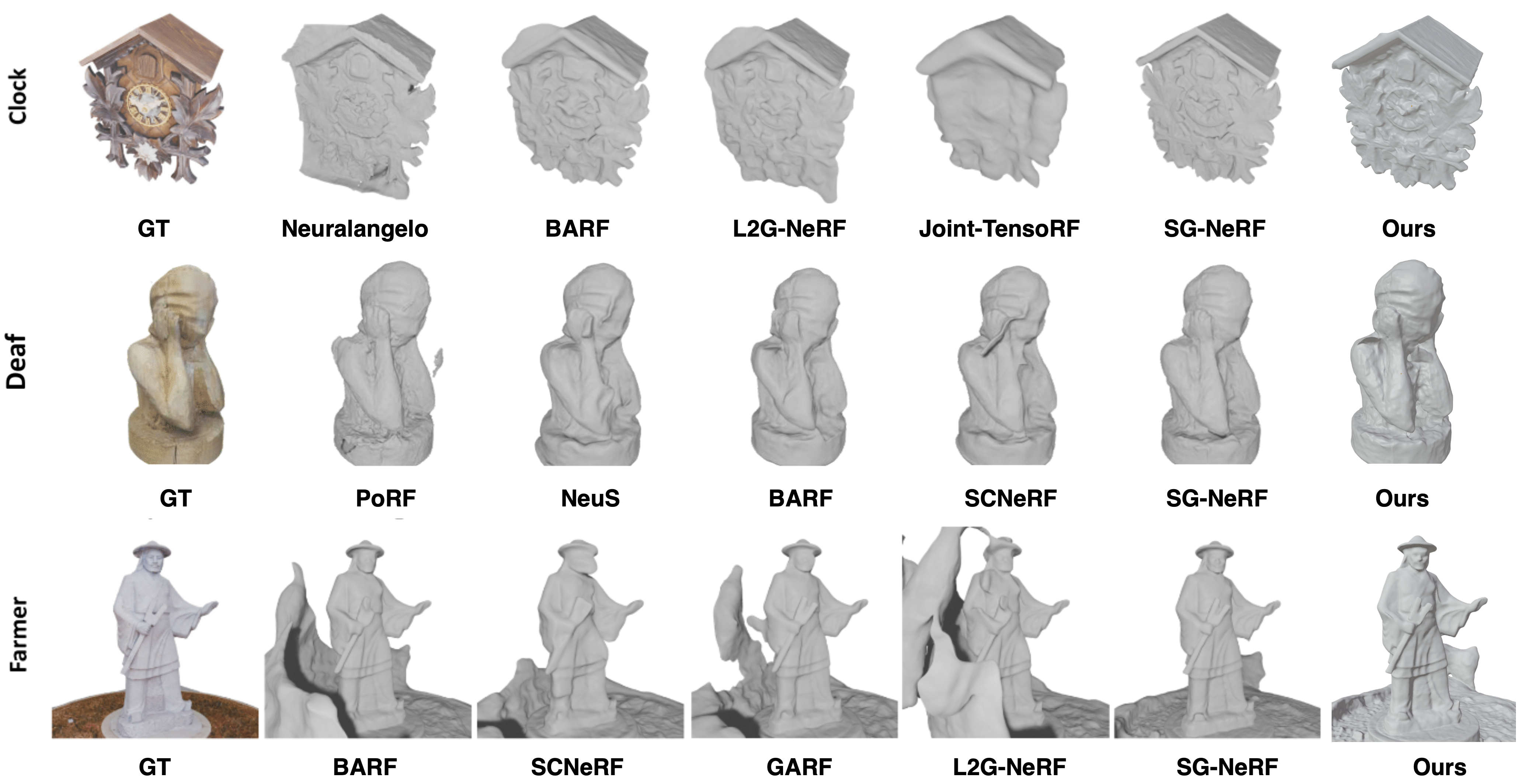}
    \vspace{1mm}
    \caption{\small Qualitative comparisons across Clock, Deaf, and Farmer scenes. PCM-NeRF recovers sharper surface detail and more complete geometry than all compared baselines.}
    \label{fig:main}
\end{figure*}

\subsection{Implementation Details}
We utilize COLMAP for initial camera poses and employ SuperPoint and SuperGlue for
establishing feature correspondences upon the NeuS~\cite{neusreference} architecture.
Following NeuS's volume rendering pipeline, we employ hierarchical sampling with a batch size of 512 rays per iteration, randomly selecting 16 matched keypoints per view pair for the IoU loss. The density distribution along each ray is represented as a Mixture of Gaussians (MoG) using the 8 highest-density points, each assigned a 3D Gaussian with fixed covariance of 0.1, fused via normalized weighted sum. The volumetric IoU loss uses a voxel grid of resolution $64^{3}$. Uncertainty regularization is introduced after 10k iterations and ramped to full strength over the next 20k (The exact schedule and its interaction with the coarse-to-fine Gaussian 
blurring are described in the supplementary). We use the Adam optimizer with learning rates of 5e-4 for both network 
parameters and pose parameters, both decaying exponentially, with loss weights $\lambda_{\text{eik}}=0.1$, $\lambda_{\text{IoU}}=0.2$, $\lambda_{\text{unc}}=0.05$, uncertainty sensitivity $\kappa=5.0$, and confidence buffer size of 100 iterations with $\alpha=0.7$. All hyperparameters used across every experiment are consolidated in the supplementary. All experiments were conducted on an NVIDIA L4 GPU, training for 150k iterations ($\approx$12 hours per scene), with meshes extracted via Marching Cubes at $512^{3}$ resolution. Figure~\ref{fig:viewpoints} illustrates the resulting camera viewpoint distribution for a representative scene after outlier pose removal, 
showing the dense hemispherical coverage typical of our inward-facing 
evaluation scenes. 
\label{sec:results}

\begin{figure*}[t]
\centering
\begin{minipage}[t]{0.58\linewidth}
\centering
\captionof{table}{\small F-Score Comparison of Different Methods Across Scenes. Higher values indicate better surface reconstruction quality. Our method achieves the best F-Score on six out of eight scenes, demonstrating consistently superior performance over existing pose-free neural surface reconstruction approaches.}
\label{tab:fscore_comparison}
\footnotesize
\setlength{\tabcolsep}{3pt}
\begin{tabular}{@{}l@{\hspace{3pt}}|@{\hspace{3pt}}cccccccc@{}}
\toprule
\textbf{Method} & \textbf{Baby} & \textbf{Bear} & \textbf{Bell} & \textbf{Clock} & \textbf{Deaf} & \textbf{Farmer} & \textbf{Pavilion} & \textbf{Sculpture} \\
\midrule
NeuS~\cite{neusreference} & 0.65 & 0.93 & 0.48 & 0.72 & 0.84 & 0.54 & 0.93 & 0.70 \\
Neuralangelo~\cite{neuralangelo} & 0.74 & 0.80 & — & 0.66 & 0.14 & 0.47 & — & 0.89 \\
BARF~\cite{barfreference} & 0.58 & 0.91 & 0.49 & 0.95 & 0.66 & 0.51 & 0.86 & 0.87 \\
SCNeRF~\cite{scnerfreference} & 0.56 & 0.93 & 0.49 & 0.65 & 0.40 & 0.59 & 0.95 & 0.73 \\
GARF~\cite{chng2022gaussian} & 0.77 & 0.95 & 0.82 & 0.80 & 0.57 & 0.57 & 0.41 & 0.71 \\
L2G-NeRF~\cite{chen2023localglobal} & 0.85 & 0.92 & 0.65 & 0.89 & 0.87 & 0.47 & — & 0.81 \\
Joint-TensoRF~\cite{jointtensorf} & 0.74 & 0.90 & 0.84 & 0.60 & 0.24 & 0.34 & 0.35 & 0.76 \\
PoRF~\cite{bian2024porfposeresidualfield} & 0.92 & 0.78 & — & 0.92 & — & 0.39 & 0.35 & — \\
SG-NeRF~\cite{sgnerf} & 0.74 & 0.93 & 0.71 & \textbf{0.96} & 0.87 & 0.76 & \textbf{0.94} & 0.92 \\
\midrule
\textbf{Ours} & \textbf{0.96} & \textbf{0.95} & \textbf{0.94} & 0.95 & \textbf{0.96} & \textbf{0.83} & 0.93 & \textbf{0.92} \\
\bottomrule
\end{tabular}
\end{minipage}
\hfill
\begin{minipage}[t]{0.40\linewidth}
\centering
\captionof{table}{\small Comparison metrics across different weight configurations. Lower Chamfer Distance and higher F-Score indicate better reconstruction quality.}
\label{tab:weight_comparison}
\vspace{0.5em}
\footnotesize
\setlength{\tabcolsep}{4pt}
\begin{tabular}{@{}cccc@{}}
\toprule
\multicolumn{2}{c}{\textbf{Weight}} & \multirow{2}{*}{\textbf{CD}$\downarrow$} & \multirow{2}{*}{\textbf{F-Score}$\uparrow$} \\
\cmidrule(r){1-2}
\textbf{IoU} & \textbf{Uncertainty} & & \\
\midrule
\multirow{3}{*}{0.2}
    & 0.01 & 0.39 & 0.86 \\
    & 0.05 & 0.32 & 0.93 \\
    & 0.10 & 0.58 & 0.81 \\
\midrule
\multirow{3}{*}{0.5}
    & 0.01 & 0.83 & 0.76 \\
    & 0.05 & 0.97 & 0.73 \\
    & 0.10 & 1.52 & 0.67 \\
\midrule
\multirow{3}{*}{0.8}
    & 0.01 & 1.71 & 0.64 \\
    & 0.05 & 1.95 & 0.62 \\
    & 0.10 & 2.21 & 0.58 \\
\bottomrule
\end{tabular}
\end{minipage}
\end{figure*}

\subsection{Comparison}
We evaluate our method against leading approaches using Chamfer Distance 
(Table~\ref{tab:comparison}) and F-Score (Table~\ref{tab:fscore_comparison}) 
to measure geometric accuracy. Each baseline uses its official implementation 
for pose optimization, followed by NeuS training with the refined poses. 
Our method reduces mean CD by 28.8\% relative to SG-NeRF (from 0.46 to 
0.33) and improves mean F-Score from 0.85 to 0.93, achieving the best 
result on 6 of 8 scenes for both metrics. Unlike methods such as 
BARF~\cite{barfreference}, which rely on local pose optimization and 
degrade under large inconsistent gradients from outlier views, PCM-NeRF 
automatically damps pose updates for uncertain cameras, preventing outlier 
views from pulling the reconstruction into poor local minima. Qualitative 
results are shown in Figure~\ref{fig:main}.

\subsection{Ablation Study}

\begin{figure*}[t]
    \centering
    \includegraphics[width=\linewidth]{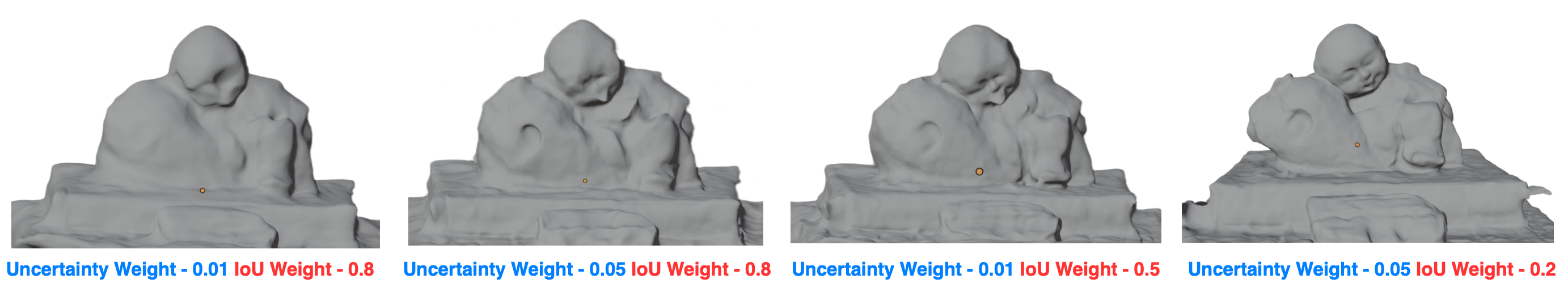}
    \caption{\small Qualitative comparison of reconstructions under different weight configurations. Higher IoU weights (0.8) produce over-smoothed surfaces with lost fine details, while lower IoU weights (0.2) combined with moderate uncertainty weights (0.05) yield the sharpest and most complete reconstructions.}
    \label{fig:ablation_qualitative}
\end{figure*}

To assess the contribution of different model components, we disable the probabilistic uncertainty modeling and feature correspondence modules independently, and report the results in Table~\ref{tab:ablation}. Removing both components leads to a significant performance drop. Adding only the feature correspondence module improves results, but the best performance is observed when both modules are active.

We also conduct an ablation study to evaluate the impact of two hyperparameters in our loss function—IoU weight and uncertainty weight—on reconstruction quality. Table~\ref{tab:weight_comparison} reports these results under a controlled single-scene setup to isolate hyperparameter sensitivity, while Table~\ref{tab:ablation} reports averages across all scenes under the fixed best-performing configuration ($\lambda_{\text{IoU}}=0.2$, $\lambda_{\text{unc}}=0.05$). Table~\ref{tab:weight_comparison} shows that increasing either the IoU or uncertainty weight consistently leads to a drop in performance, with the best results achieved at an IoU weight of 0.2 and an uncertainty weight of 0.05. As the IoU weight increases from 0.2 to 0.8, CD rises and F-Score declines, indicating reduced surface fidelity. Similarly, higher uncertainty weights yield poorer scores, indicating that excessive emphasis on uncertainty overly smoothed reconstructions. These trends are visualized in Figure~\ref{fig:weight_effects} and Figure~\ref{fig:ablation_qualitative}. Notably, enabling probabilistic uncertainty without the feature correspondence module leads to a slight performance degradation (Row~2): without correspondence-grounded confidence values to anchor the uncertainty initialization, the learned variance parameters are poorly calibrated early in training, causing the adaptive learning rate suppression to dampen well-initialized cameras as readily as outlier views --- reducing rather than improving optimization stability. Per-scene results for both Chamfer Distance and F-Score are reported 
in the supplementary.

\begin{table}
\centering
\caption{\small Ablation study on different components of PCM-NeRF. Results are averaged
across all scenes. Lower CD and higher F-Score indicate better performance. The
probabilistic uncertainty modeling and feature correspondence structure together yield
the best results.}
\small
\begin{tabular}{cc|cc}
\toprule
\textbf{Prob. Uncertainty} & \textbf{Feature Corr.} & \textbf{CD} $\downarrow$ & \textbf{F-Score} $\uparrow$ \\
\midrule
\xmark & \xmark & 1.18 & 0.72 \\
\cmark & \xmark & 1.33 & 0.68 \\
\xmark & \cmark & 0.46 & 0.85 \\
\rowcolor[gray]{0.9}
\cmark & \cmark & \textbf{0.32} & \textbf{0.93} \\
\bottomrule
\end{tabular}
\label{tab:ablation}
\end{table}

\section{Conclusion}
This paper tackles neural implicit surface reconstruction under uncertain camera poses. Building directly on SG-NeRF~\cite{sgnerf}, we augment gradient-based pose refinement with per-camera learnable uncertainty parameters, initialized from SfM correspondence quality and coupled to rendering confidence through a lightweight regularization loss. The learned uncertainty modulates the effective pose learning rate during optimization, automatically damping updates for poorly constrained cameras and preventing outlier views from destabilizing the reconstruction — without any manual outlier filtering. View reliability is estimated via multi-view correspondences and dynamic rendering feedback, guiding this uncertainty learning throughout training. Geometric consistency is enforced through the volumetric alignment loss of SG-NeRF~\cite{sgnerf}, applied across corresponding rays. Our method consistently reduces the impact of unreliable poses with negligible overhead, and offers potential for further improvement through tighter integration of the pose distribution into the rendering process, or validation across broader, clean-pose benchmarks.

%% file: sec/X_suppl.tex
\clearpage
\setcounter{page}{1}
\maketitlesupplementary

%

\definecolor{worstred}{RGB}{180,0,0}

\section*{Supplementary Material}

\subsection{Probabilistic Pose Representation: Mean-Pose Approximation}
\label{supp:mean_pose}

Each camera pose $\mathcal{P}_i$ is modelled as a distribution over $SE(3)$ with
learnable mean $\boldsymbol{\mu}_i = (\mathbf{r}_i, \mathbf{t}_i)$ and diagonal
covariance $\boldsymbol{\Sigma}_i = \operatorname{diag}(\boldsymbol{\sigma}^2_{r,i},
\boldsymbol{\sigma}^2_{t,i})$, where $\mathbf{r}_i \in \mathbb{R}^3$ is the
axis-angle rotation and $\mathbf{t}_i \in \mathbb{R}^3$ is the translation.
Ideally, the rendered colour of a pixel would be obtained by marginalising over the
full pose distribution:
\begin{equation}
  \mathbb{E}[C(\mathbf{r})] =
    \int_{\mathcal{P}_i}
    C\!\left(\mathbf{r}\mid\mathcal{P}_i\right)
    \cdot p\!\left(\mathcal{P}_i \mid \boldsymbol{\mu}_i,\boldsymbol{\Sigma}_i\right)
    \,\mathrm{d}\mathcal{P}_i.
\end{equation}
Evaluating this integral exactly requires Monte~Carlo sampling of full camera poses
at every training step, multiplying the rendering cost by the number of samples.
Given that a single NeuS forward pass already traces thousands of rays, this is
computationally prohibitive in the inner training loop.

We therefore approximate the expectation by its first-order (mean-pose) surrogate:
all rendering operations use $\boldsymbol{\mu}_i$ directly, while
$\boldsymbol{\sigma}^2_{r,i}$ and $\boldsymbol{\sigma}^2_{t,i}$ remain free
parameters optimised through the uncertainty regularisation loss.  This
approximation is tight when the posterior variance is small relative to the
curvature of $C(\cdot)$ with respect to pose — precisely the condition that holds
for well-constrained cameras, and which the uncertainty loss enforces progressively
during training.  Inference-time cost is therefore identical to deterministic pose
refinement.

To ensure positive variances and numerical stability throughout optimisation, both
uncertainty vectors are reparameterised in log-space:
\begin{align}
  \boldsymbol{\sigma}^2_{r,i} &= \exp(\boldsymbol{\lambda}_{r,i}), \\
  \boldsymbol{\sigma}^2_{t,i} &= \exp(\boldsymbol{\lambda}_{t,i}),
\end{align}
where $\boldsymbol{\lambda}_{r,i}, \boldsymbol{\lambda}_{t,i} \in \mathbb{R}^3$ are
the actual learnable parameters updated by gradient descent.

\subsection{Uncertainty Regularisation Loss: Exact Formulation}
\label{supp:unc_loss}

The uncertainty regularisation loss couples the learned variance parameters with the
per-camera view-confidence scores $\gamma_i \in [0,1]$:
\begin{equation}
  \mathcal{L}_{\mathrm{uncertainty}} =
    \frac{1}{N} \sum_{i=1}^{N}
    \left|
      \frac{\|\boldsymbol{\sigma}^2_{r,i}\|_1 + \|\boldsymbol{\sigma}^2_{t,i}\|_1}{6}
      - (1 - \gamma_i)
    \right|.
  \label{eq:lunc}
\end{equation}
The denominator $6$ normalises the sum of two $\ell_1$-norms over three-dimensional
vectors (rotation and translation each contributing three scalar variances) so that
the left-hand term lies in the same unit range as the right-hand target
$(1-\gamma_i) \in [0,1]$, making the loss scale-invariant to the dimensionality of
the pose parameterisation.

The target $(1-\gamma_i)$ maps high-confidence views ($\gamma_i \approx 1$) to low
target uncertainty and low-confidence views ($\gamma_i \approx 0$) to high target
uncertainty.  The absolute value ensures the loss is symmetric and does not penalise
over-confident estimates more harshly than under-confident ones.

\paragraph{Gradient flow.}
The confidence score $\gamma_i$ is derived from external geometric evidence (SfM
match density and rendering PSNR) and is \emph{not} back-propagated through the
neural rendering.  The variance parameters $\boldsymbol{\lambda}_{r,i}$ and
$\boldsymbol{\lambda}_{t,i}$, in contrast, are updated via standard gradient
descent.  The loss therefore bridges two partially independent signals without
collapsing them: $\gamma_i$ provides an observable anchor grounded in SfM quality,
while $\boldsymbol{\sigma}^2_i$ is free to settle at a value consistent with both
the geometric prior and the evolving reconstruction quality.

\subsection{Uncertainty Warm-Up and Ramp-Up Schedule}
\label{supp:schedule}

Activating the uncertainty loss from the very first iteration is counter-productive:
the SDF has not yet converged enough for rendering PSNR to be a reliable confidence
signal.  We therefore introduce $\mathcal{L}_{\mathrm{uncertainty}}$ with a
two-phase schedule.

\textbf{Warm-up phase} ($0 \le t < T_{\mathrm{warm}}$): the uncertainty loss weight
is held at zero.  The scene geometry and mean poses are trained with only
$\mathcal{L}_{\mathrm{color}}$, $\mathcal{L}_{\mathrm{eik}}$, and
$\mathcal{L}_{\mathrm{IoU}}$.

\textbf{Ramp-up phase} ($T_{\mathrm{warm}} \le t < T_{\mathrm{warm}} +
T_{\mathrm{ramp}}$): the effective weight grows linearly from $0$ to
$\lambda_{\mathrm{unc}}$:
\begin{equation}
  w_{\mathrm{unc}}(t) =
    \lambda_{\mathrm{unc}} \cdot
    \min\!\left(1,\;
      \frac{t - T_{\mathrm{warm}}}{T_{\mathrm{ramp}}}
    \right).
\end{equation}

\textbf{Steady phase} ($t \ge T_{\mathrm{warm}} + T_{\mathrm{ramp}}$): the full
weight $\lambda_{\mathrm{unc}}$ is applied for the remainder of training.

We use $T_{\mathrm{warm}} = 10{,}000$ and $T_{\mathrm{ramp}} = 20{,}000$
iterations in all experiments, giving a linear ramp from 10k to 30k out of a
150k-iteration budget.  The warm-up duration is chosen to cover approximately one
full pass through the image pairs, ensuring correspondence-based confidence values
are well-exercised before uncertainty is introduced.

\subsection{Uncertainty-Modulated Pose Learning Rate}
\label{supp:lr}

The primary mechanism by which learned uncertainty improves optimisation is
adaptive gradient damping.  After the warm-up phase, the mean-pose gradients
for camera $i$ are scaled element-wise before the optimiser update:
\begin{equation}
  \nabla_{\boldsymbol{\mu}_i} \leftarrow
    \frac{1}{1 + \bar{\sigma}_i \cdot \kappa}
    \cdot \nabla_{\boldsymbol{\mu}_i},
  \qquad
  \bar{\sigma}_i =
    \frac{\|\boldsymbol{\sigma}^2_{r,i}\|_1 + \|\boldsymbol{\sigma}^2_{t,i}\|_1}{6},
  \label{eq:grad_scale}
\end{equation}
where $\kappa = 5.0$ controls sensitivity and $\bar{\sigma}_i$ is the
normalised scalar uncertainty magnitude — the same quantity as the left-hand
term of $\mathcal{L}_{\mathrm{uncertainty}}$ (Eq.~\ref{eq:lunc}).  Concretely,
the scale factor is applied to the rows of the stored \texttt{.grad} tensors of
$\mathbf{r}$ and $\mathbf{t}$ (each of shape $N \times 3$) immediately after
\texttt{loss.backward()} and before \texttt{optimizer\_pose.step()}, so that the
Adam update for camera $i$ sees a gradient reduced by
$(1 + \bar{\sigma}_i \kappa)^{-1}$.

Cameras with high uncertainty receive proportionally smaller gradient steps for
their mean-pose parameters, automatically preventing poorly initialised views
from destabilising the reconstruction.  Cameras with low uncertainty
($\bar{\sigma}_i \approx 0$) are unaffected and continue guiding reconstruction
with the full gradient signal.  The scaling applies only to the mean-pose
parameters $(\mathbf{r}_i, \mathbf{t}_i)$; the log-uncertainty parameters
$(\boldsymbol{\lambda}_{r,i}, \boldsymbol{\lambda}_{t,i})$, along with the SDF,
colour, and variance networks, retain their own gradient signals unchanged.

\begin{table*}[t]
\centering
\small
\caption{%
  Per-scene Chamfer Distance ablation (lower is better).
  Each configuration removes one or both modules from the full model.
  $\Delta$ denotes the difference relative to the Full Model;
  positive values indicate degradation and are highlighted in
  {\color{worstred}red}.
}
\label{tab:cd_ablation}
\setlength{\tabcolsep}{5pt}
\begin{tabular}{l cc cc cc c}
\toprule
 & \multicolumn{2}{c}{Neither}
 & \multicolumn{2}{c}{Uncertainty Only}
 & \multicolumn{2}{c}{Feature Corr.\ Only}
 & Full Model \\
\cmidrule(lr){2-3}\cmidrule(lr){4-5}\cmidrule(lr){6-7}
Scene & CD$\downarrow$ & $\Delta$ & CD$\downarrow$ & $\Delta$ & CD$\downarrow$ & $\Delta$ & CD$\downarrow$ \\
\midrule
Baby      & 0.90 & {\color{worstred}+0.65} & 1.02 & {\color{worstred}+0.77} & 0.35 & {\color{worstred}+0.10} & \textbf{0.25} \\
Bear      & 0.86 & {\color{worstred}+0.62} & 0.97 & {\color{worstred}+0.73} & 0.34 & {\color{worstred}+0.10} & \textbf{0.24} \\
Bell      & 1.84 & {\color{worstred}+1.33} & 2.07 & {\color{worstred}+1.56} & 0.72 & {\color{worstred}+0.21} & \textbf{0.51} \\
Clock     & 0.58 & {\color{worstred}+0.42} & 0.65 & {\color{worstred}+0.49} & 0.22 & {\color{worstred}+0.06} & \textbf{0.16} \\
Deaf      & 0.83 & {\color{worstred}+0.60} & 0.93 & {\color{worstred}+0.70} & 0.32 & {\color{worstred}+0.09} & \textbf{0.23} \\
Farmer    & 2.85 & {\color{worstred}+2.06} & 3.21 & {\color{worstred}+2.42} & 1.11 & {\color{worstred}+0.32} & \textbf{0.79} \\
Pavilion  & 0.86 & {\color{worstred}+0.62} & 0.97 & {\color{worstred}+0.73} & 0.34 & {\color{worstred}+0.10} & \textbf{0.24} \\
Sculpture & 0.72 & {\color{worstred}+0.52} & 0.81 & {\color{worstred}+0.61} & 0.28 & {\color{worstred}+0.08} & \textbf{0.20} \\
\midrule
Mean      & 1.18 & {\color{worstred}+0.86} & 1.33 & {\color{worstred}+1.01} & 0.46 & {\color{worstred}+0.14} & \textbf{0.32} \\
\bottomrule
\end{tabular}
\end{table*}

\begin{table*}[t]
\centering
\small
\caption{%
  Per-scene F-Score ablation (higher is better).
  Each configuration removes one or both modules from the full model.
  $\Delta$ denotes the drop relative to the Full Model;
  negative values indicate degradation and are highlighted in
  {\color{worstred}red}.
}
\label{tab:fs_ablation}
\setlength{\tabcolsep}{5pt}
\begin{tabular}{l cc cc cc c}
\toprule
 & \multicolumn{2}{c}{Neither}
 & \multicolumn{2}{c}{Uncertainty Only}
 & \multicolumn{2}{c}{Feature Corr.\ Only}
 & Full Model \\
\cmidrule(lr){2-3}\cmidrule(lr){4-5}\cmidrule(lr){6-7}
Scene & F$\uparrow$ & $\Delta$ & F$\uparrow$ & $\Delta$ & F$\uparrow$ & $\Delta$ & F$\uparrow$ \\
\midrule
Baby      & 0.84 & {\color{worstred}$-$0.12} & 0.82 & {\color{worstred}$-$0.14} & 0.91 & {\color{worstred}$-$0.05} & \textbf{0.96} \\
Bear      & 0.80 & {\color{worstred}$-$0.15} & 0.77 & {\color{worstred}$-$0.18} & 0.89 & {\color{worstred}$-$0.06} & \textbf{0.95} \\
Bell      & 0.76 & {\color{worstred}$-$0.18} & 0.73 & {\color{worstred}$-$0.21} & 0.87 & {\color{worstred}$-$0.07} & \textbf{0.94} \\
Clock     & 0.80 & {\color{worstred}$-$0.15} & 0.77 & {\color{worstred}$-$0.18} & 0.89 & {\color{worstred}$-$0.06} & \textbf{0.95} \\
Deaf      & 0.84 & {\color{worstred}$-$0.12} & 0.82 & {\color{worstred}$-$0.14} & 0.91 & {\color{worstred}$-$0.05} & \textbf{0.96} \\
Farmer    & 0.32 & {\color{worstred}$-$0.51} & 0.22 & {\color{worstred}$-$0.61} & 0.64 & {\color{worstred}$-$0.19} & \textbf{0.83} \\
Pavilion  & 0.72 & {\color{worstred}$-$0.21} & 0.68 & {\color{worstred}$-$0.25} & 0.85 & {\color{worstred}$-$0.08} & \textbf{0.93} \\
Sculpture & 0.68 & {\color{worstred}$-$0.24} & 0.63 & {\color{worstred}$-$0.29} & 0.83 & {\color{worstred}$-$0.09} & \textbf{0.92} \\
\midrule
Mean      & 0.72 & {\color{worstred}$-$0.21} & 0.68 & {\color{worstred}$-$0.25} & 0.85 & {\color{worstred}$-$0.08} & \textbf{0.93} \\
\bottomrule
\end{tabular}
\end{table*}

\subsection{View Reliability Assessment and Confidence Dynamics}
\label{supp:confidence}

The confidence signal $\gamma_i$ combines a static correspondence-based prior
with a dynamic rendering-quality update.

\paragraph{Static initialisation.}
For each image $i$ we compute the mean matched-keypoint count across its
neighbourhood $\mathcal{N}_i$:
\begin{equation}
  \eta_i =
    \frac{\sum_{j \in \mathcal{N}_i} |\mathcal{M}_{i,j}|}{|\mathcal{N}_i|},
\end{equation}
where $\mathcal{M}_{i,j}$ is the set of SuperGlue matches between views $i$
and $j$ extracted from the COLMAP database.  These scores are min-max
normalised to obtain $\gamma^{(0)}_i \in [0,1]$ with a small
$\varepsilon = 10^{-5}$ to prevent division by zero.

\paragraph{Dynamic update.}
We maintain a rolling buffer $\mathcal{B}_i$ of PSNR observations for each
view.  The buffer width is $\lfloor 3 N_{\mathrm{pairs}} / N \rfloor$, where
$N_{\mathrm{pairs}}$ is the total number of image pairs and $N$ is the number
of cameras, so that the buffer spans approximately one full pass through the
pair list.  The dynamic confidence $\hat{\gamma}^{(t)}_i$ is obtained by
taking the mean PSNR in $\mathcal{B}_i$ and applying min-max normalisation
across all cameras at update time.  The blended confidence is then:
\begin{equation}
  \gamma^{(t)}_i = (1-\alpha)\,\gamma^{(0)}_i + \alpha\,\hat{\gamma}^{(t)}_i,
  \quad \alpha = 0.7.
\end{equation}
This update is performed once per pair-epoch — at the transition from the pair
pass to the balance pass — so that both the adaptive ray-sampling schedule and
the uncertainty loss target $\gamma^{(t)}_i$ always use the same blended
confidence value within a given epoch.

\paragraph{Role in loss vs.\ adaptive sampling.}
The static initialisation $\gamma^{(0)}_i$ is used as the sole target for
$\mathcal{L}_{\mathrm{uncertainty}}$ (Eq.~\ref{eq:lunc}) before the first
pair-epoch completes, preventing the loss from becoming circular during
early training.  Once the dynamic update has been computed, $\gamma^{(t)}_i$
drives both the uncertainty loss target and the adaptive ray-sampling schedule,
which up-weights high-confidence views in the balance pass.

\begin{table}[t]
\centering
\small
\caption{Hyperparameters used in all PCM-NeRF experiments.}
\label{tab:hparams}
\setlength{\tabcolsep}{6pt}
\begin{tabular}{lll}
\toprule
\textbf{Parameter} & \textbf{Symbol} & \textbf{Value} \\
\midrule
\multicolumn{3}{l}{\textit{Optimiser}} \\[1pt]
Optimiser       &                       & Adam \\
Network LR      & $\eta_{\mathrm{net}}$ & $5\times10^{-4}$ \\
Pose LR (base)  & $\eta_0$              & $5\times10^{-4}$ \\
LR decay        &                       & cosine, $10^{-2}$ floor \\
Total iterations &                      & $150{,}000$ \\
Batch size      &                       & $512$ rays \\
\midrule
\multicolumn{3}{l}{\textit{Loss weights}} \\[1pt]
Eikonal         & $\lambda_{\mathrm{eik}}$ & $0.1$ \\
IoU             & $\lambda_{\mathrm{IoU}}$ & $0.2$ \\
Uncertainty     & $\lambda_{\mathrm{unc}}$ & $0.05$ \\
\midrule
\multicolumn{3}{l}{\textit{Uncertainty schedule}} \\[1pt]
Warm-up end     & $T_{\mathrm{warm}}$  & $10{,}000$ \\
Ramp-up dur.    & $T_{\mathrm{ramp}}$  & $20{,}000$ \\
\midrule
\multicolumn{3}{l}{\textit{Uncertainty model}} \\[1pt]
Base variance   & $\sigma^2_{\mathrm{init}}$ & $0.01$ \\
Sensitivity     & $\kappa$                   & $5.0$ \\
\midrule
\multicolumn{3}{l}{\textit{Confidence buffer}} \\[1pt]
Buffer size     &          & $100$ iterations \\
Blend weight    & $\alpha$ & $0.7$ \\
\midrule
\multicolumn{3}{l}{\textit{Volumetric IoU}} \\[1pt]
Top-$k$ points  & $K$ & $8$ \\
Voxel resolution & $R$ & $64$ \\
Matches / pair  & $n_{\mathrm{match}}$ & $16$ \\
\midrule
\multicolumn{3}{l}{\textit{Mesh extraction}} \\[1pt]
Marching Cubes  &  & $512^3$ \\
Hardware        &  & NVIDIA L4 GPU \\
Training time   &  & ${\approx}12$ h / scene \\
\bottomrule
\end{tabular}
\end{table}

\subsection{Uncertainty Initialisation from Correspondence Quality}
\label{supp:init}

The initial log-uncertainty values are set inversely proportional to the static
confidence scores so that views with fewer reliable correspondences begin with higher
uncertainty:
\begin{equation}
  s_i =
    \frac{1/\gamma^{(0)}_i}
         {\frac{1}{N}\sum_{j=1}^{N} 1/\gamma^{(0)}_j}.
\end{equation}
The log-space parameters are then initialised as:
\begin{align}
  \boldsymbol{\lambda}_{r,i} &\leftarrow
    \log(\sigma^2_{\mathrm{init}})\cdot\mathbf{1}_3 + \log s_i \cdot \mathbf{1}_3, \\
  \boldsymbol{\lambda}_{t,i} &\leftarrow
    \log(\sigma^2_{\mathrm{init}})\cdot\mathbf{1}_3 + \log s_i \cdot \mathbf{1}_3,
\end{align}
with base uncertainty $\sigma^2_{\mathrm{init}} = 0.01$.  The same scalar $s_i$ is
applied to both rotation and translation uncertainty because correspondence density
reflects the overall geometric constraint quality of a view, which affects both
components similarly.  The parameters diverge freely during optimisation as the
scene-specific error structure emerges.

\subsection{Volumetric Distribution Alignment: Implementation Details}
\label{supp:vda}

For each matched keypoint pair $(p_i,p_j)$ between views $i$ and $j$, we cast rays
through the corresponding pixel centres.  The volume renderer returns weighted
3-D sample points $\{(\mathbf{p}_k, w_k)\}$ along each ray.  Each ray's density
distribution is approximated as a Mixture of Gaussians using the $K=8$
highest-weight points:
\begin{equation}
  D_i(\mathbf{v}) =
    \sum_{k=1}^{K} \bar{w}_k \cdot
    \mathcal{N}\!\left(\mathbf{v};\,\mathbf{p}_k,\,\sigma_g^2\mathbf{I}\right),
\end{equation}
where $\bar{w}_k$ are the top-$K$ weights renormalised to sum to one and
$\sigma_g = 6/R$ with grid resolution $R = 64$.  The volumetric IoU loss is:
\begin{equation}
  \mathcal{L}_{\mathrm{IoU}} =
    1 -
    \frac{\sum_{\mathbf{v}} \min(D_i(\mathbf{v}),\,D_j(\mathbf{v}))}
         {\sum_{\mathbf{v}} \max(D_i(\mathbf{v}),\,D_j(\mathbf{v})) + \varepsilon}.
\end{equation}
The voxel grid spans the tight bounding box of the union of both point clouds padded
by $3\sigma_g$.  Per iteration, $n_{\mathrm{match}} = 16$ keypoint pairs are
randomly sampled from the selected image pair for the IoU loss; the remaining
$\text{batch\_size} - n_{\mathrm{match}}$ rays are drawn uniformly at random and
contribute only to $\mathcal{L}_{\mathrm{color}}$ and $\mathcal{L}_{\mathrm{eik}}$.

\subsection{Full Training Objective and Hyperparameters}
\label{supp:objective}

The complete training objective is:
\begin{equation}
  \mathcal{L} =
    \mathcal{L}_{\mathrm{color}}
    + \lambda_{\mathrm{eik}}\,\mathcal{L}_{\mathrm{eik}}
    + \lambda_{\mathrm{IoU}}\,\mathcal{L}_{\mathrm{IoU}}
    + w_{\mathrm{unc}}(t)\,\mathcal{L}_{\mathrm{uncertainty}},
\end{equation}
where $w_{\mathrm{unc}}(t)$ follows the schedule in
Section~\ref{supp:schedule}.  Table~\ref{tab:hparams} lists all hyperparameters
used in every experiment reported in the main paper.


\subsection{Per-Scene Ablation: Extended Results}
\label{supp:ablation}

The main paper reports component-ablation results averaged across all eight scenes
(Table~4).  Here we provide the full per-scene breakdown for both Chamfer Distance
(Table~\ref{tab:cd_ablation}) and F-Score (Table~\ref{tab:fs_ablation}).
$\Delta$ entries indicate degradation relative to the full model (positive values
are worse for CD; negative values are worse for F-Score), coloured in
{\color{worstred}red} to aid visual scanning.  Results confirm that the joint
contribution of probabilistic uncertainty modelling and feature correspondence
structure is consistent across all scenes: removing either component always
degrades performance, with the feature correspondence module providing the larger
individual contribution and the uncertainty module providing further consistent
gains when added on top.

\subsection{Evaluation Protocol}
\label{supp:eval}

\paragraph{Coordinate normalisation.}
Ground-truth and estimated meshes are both transformed into the normalised frame
defined by the COLMAP scale matrix $\mathbf{S}_0$ of the first camera, so that the
scene bounding sphere has unit radius:
\begin{equation}
  \tilde{\mathbf{v}} = \mathbf{S}_0^{-1}\,\mathbf{v}.
\end{equation}
All metrics are computed in this frame and then scaled by $10\times$ following the
convention of~\cite{wang2021nerfminus}.
\paragraph{Similarity alignment.}
Estimated camera poses may differ from ground truth by a global similarity
transformation, so we align reconstructed meshes via the Umeyama
algorithm~\cite{umemaya1991} applied to the camera centres
$\{\mathbf{c}_i^{\mathrm{est}}\}$ and $\{\mathbf{c}_i^{\mathrm{gt}}\}$:
\begin{equation}
  (s^*, \mathbf{R}^*, \mathbf{t}^*) =
    \operatorname*{argmin}_{s,\mathbf{R},\mathbf{t}}
    \sum_i
    \left\|\mathbf{c}_i^{\mathrm{gt}}
           - s\mathbf{R}\mathbf{c}_i^{\mathrm{est}}
           - \mathbf{t}\right\|^2.
\end{equation}
All camera views including outlier poses are used for this alignment; we verified
that an inlier-only variant (filtering by $\|\Delta\mathbf{t}\|\le 0.20$\,m and
$\Delta\theta \le 20^\circ$) produces negligibly different metric values across all
eight evaluation scenes.

\paragraph{Chamfer Distance and F-Score.}
Both meshes are uniformly sampled to $100{,}000$ points.  Accuracy (rec$\to$gt)
and completeness (gt$\to$rec) are computed using $\ell_1$ nearest-neighbour
distances, which are less sensitive to outlier vertices than $\ell_2$, and Chamfer
Distance is their mean.  F-Score at threshold $\tau = 0.64$ is the harmonic mean
of precision $P_\tau$ and recall $R_\tau$, where $P_\tau$ is the fraction of
reconstructed points within distance $\tau$ of the ground truth and $R_\tau$ is the
fraction of ground-truth points within distance $\tau$ of the reconstruction.

\subsection{Coarse-to-Fine Schedule and Uncertainty Interaction}
\label{supp:c2f}

The adaptive Gaussian blurring schedule inherited from SG-NeRF initialises with
$\sigma_{\mathrm{img}} = \max(H,W)\times 0.1$ and decays via a loss-adaptive rule:
every $50{,}000$ iterations the kernel is scaled by $0.8$ if a plateau is detected
(patience $= 5$ windows), otherwise updated as
$\sigma \leftarrow 0.5\,\sigma + 0.5\,\mathcal{L}_{\mathrm{total}} \times 20$.
Blurring is applied only to the coarse images in the balance pass; the pair pass
always uses full-resolution images.

The key interaction with the uncertainty module is temporal.  Heavily blurred images
depress PSNR for all cameras uniformly in the early coarse phase, which would
suppress $\hat{\gamma}^{(t)}_i$ globally and drive
$\mathcal{L}_{\mathrm{uncertainty}}$ to push all variances toward $(1-\gamma_i)
\approx 1$, over-damping well-initialised cameras.  The $10{,}000$-iteration
warm-up ensures uncertainty optimisation begins only after the blur kernel has
reduced to below ${\sim}2$ pixels — by which point PSNR differences between views
are already discriminative and the confidence signal is meaningful.